\documentclass[conference]{IEEEtran}
\IEEEoverridecommandlockouts

\usepackage{amsmath,amssymb,amsfonts}
\usepackage{algorithm}
\usepackage{algpseudocode}
\usepackage{booktabs}
\usepackage{enumitem}
\usepackage{cite}
\usepackage{array}
\usepackage{xcolor}

\title{RDEx-CASK: Cauchy Mutation, Archive, and Stagnation Kick for RDEx-CSOP}

\author{
\IEEEauthorblockN{Dikshant,\quad Dikshit Chauhan,\quad Chen Hao,\\
Anupam Trivedi,\quad K.~Harikumar,\quad J.~Senthilnath}
\IEEEauthorblockA{dikshant.gurudutt@students.iiit.ac.in,\quad dikshitchauhan608@gmail.com,\quad Chen\_Hao@a-star.edu.sg,\\
triv.anupam@gmail.com,\quad harikumar.k@iiit.ac.in,\quad J\_Senthilnath@a-star.edu.sg}
}

\begin{document}
\maketitle

\begin{abstract}
We extend RDEx-CSOP \cite{tao2026rdex} with 3 changes that target stagnation \& late-stage variance, plus minor parameter tuning. The second scale factor in the standard branch is sampled independently from a truncated Cauchy. A small feasible-only JADE-style archive ($|\mathcal{A}|_{\max} = 50$) is added \& sampled with probability $|\mathcal{A}|/(|\mathcal{A}|+|P|)$. Per-individual stagnation counter triggers, after $180$ no-improvement generations, three local overrides on standard branch: pull toward the global best, lift the archive sampling floor to $0.65$, \& saturate $CR$ to $0.95$ when population success rate is below $0.10$. The exploitation biased branch \& every other RDEx component are left untouched. On CEC CSOP suite ($D = 30$, $25$ runs), RDEx-CASK is competitive with RDEx, UDE-III, \& CL-SRDE in feasibility-aware quality \& improves time-to-target on most problems.
\end{abstract}

\section{Introduction}
RDEx-CSOP is a feasibility-aware reconstructed differential evolution variant that placed first on the CEC CSOP track \cite{tao2026rdex}. Its strong U-score performance is driven mainly by faster threshold attainment under feasibility-aware ranking rule. Two observations motivate the present extension. First, several adaptation parameters in the standard branch were tuned for an earlier benchmark cycle \& admit a small number of robust replacements that improve the speed component without disturbing the EB branch. Second, front population enters long no-improvement plateaus on many of the 28 problems in second half of the budget, after which individual targets contribute no further accepted trials. RDEx-CASK targets both observations through local changes \& leaves the EB ordered-mutation branch, the constraint handling rule, the population reduction schedule, \& success-history bookkeeping intact.

\section{Baseline RDEx-CSOP Framework}
\label{sec:baseline}
The bound-constrained problem with inequality \& equality constraints is same as in \cite{tao2026rdex}. The averaged violation measure is
\begin{equation}
\begin{aligned}
\phi(x) = \frac{1}{m_g + m_h}\!\bigg[
&\sum_{i=1}^{m_g}\!\max\{0, g_i(x)\} \\
&+ \sum_{j=1}^{m_h}\!\max\{0, |h_j(x)| - \epsilon_{eq}\}
\bigg],
\end{aligned}
\label{eq:phi}
\end{equation}
with $\epsilon_{eq} = 10^{-4}$.

Let $\mathcal{F}^{(g)} = \{x_i^{(g)}\}_{i=1}^{N^{(g)}}$ denote the front population at generation $g$. The standard branch generates a donor in the current-to-$p$ best/1 form
\begin{equation}
v_i^{(g)} = x_i^{(g)} + F_i^{(g)}\!\big(x_{pbest}^{(g)} - x_i^{(g)}\big) + F_{2,i}^{(g)}\!\big(x_{r_1}^{(g)} - x_{r_2}^{(g)}\big),
\label{eq:std}
\end{equation}
where $x_{r_1}$ is sampled from front with an exponential rank bias \&$x_{r_2}$ is sampled from current population. The EB branch uses three fitness ordered donors,
\begin{equation}
v_i^{(g)} = x_i^{(g)} + F_i^{(g)}\!\big(x_{best}^{(g)} - x_i^{(g)}\big) + F_i^{(g)}\!\big(x_{mid}^{(g)} - x_{worst}^{(g)}\big),
\label{eq:eb}
\end{equation}
\& is selected with a hybrid rate $\rho_{EB}^{(g)}$. Selection follows the $\epsilon$-rule of (9) in \cite{tao2026rdex}. The standard branch in baseline RDEx sets $F_2 = F$, samples $F$ from $\mathcal{N}\!\bigl(\max(0,(SR^{(g)})^{1/3}), 0.05^2\bigr)$, \& applies a Cauchy perturbation on non-crossover coordinates with probability $0.2$. The EB rate update sets $\rho_{EB}^{(g+1)}$ to raw improvement-share ratio when both branches contribute \& resets to $0.7$ otherwise.

\section{Proposed Variant: RDEx-CASK}
RDEx-CASK preserves (\ref{eq:eb}), the bound repair rule, the linear front reduction, the $\epsilon$-rule, \&the SHADE style memory bookkeeping. The changes are grouped into three structural modifications (Sections~\ref{sec:cauchy}, \ref{sec:archive}, \ref{sec:psk}) \& a set of parameter fine-tunings (Section~\ref{sec:fine-tunings}).

\subsection{Independent Cauchy Second Scale Factor}
\label{sec:cauchy}
In RDEx, $F_2$ is reused from $F$, so the two difference components are perfectly correlated in magnitude. We tried decoupling them, sampled $F_2$ independently from a truncated Cauchy
\begin{equation}
F_{2,i}^{(g)} \sim \mathrm{Cauchy}\bigl(\mu_F^{(g)}, 0.1\bigr) \cap [0,1],
\label{eq:F2}
\end{equation}
where $\mu_F^{(g)}$ is the centre of (\ref{eq:meanF}). The reason for picking Cauchy over Normal is the heavier tail: most draws stay near the centre, but a small fraction produce large component wise jumps, which helps escape local plateaus without permanently inflating the search variance. We tested Normal($\mu_F$, $0.1$) instead and the result was a clear regression on the speed score, which confirms the heavy tail is the load-bearing piece. We also tried scales $0.05$ \& $0.20$; both were worse than $0.10$.

\subsection{JADE-Style External Archive}
\label{sec:archive}
A circular archive $\mathcal{A}$ of capacity $|\mathcal{A}|_{\max} = 50$ is added to the standard branch. Whenever a trial replaces a target whose front entry has $\phi \le \epsilon(g)$, the displaced front entry is pushed to $\mathcal{A}$. The standard branch then samples
\begin{equation}
\tilde{x}_{r_2} =
\begin{cases}
\mathcal{A}_k,\; k \sim \mathrm{Unif}\{1,\dots,|\mathcal{A}|\}, & \text{w.p. } p_{arch}, \\
x_{r_2}, & \text{otherwise},
\end{cases}
\label{eq:archive}
\end{equation}
with
\begin{equation}
p_{arch} = \frac{|\mathcal{A}|}{|\mathcal{A}| + |P|}.
\label{eq:parch}
\end{equation}
The feasibility gate is important: an early experiment without it (push every displaced parent regardless of $\phi$) broke F06 entirely \& dragged F07 by a wide margin. Because the early phase displaced parents are mostly infeasible, \& mixing them into difference vector pulls the trial off the feasibility manifold. The capacity $|\mathcal{A}|_{\max} = 50$ was picked after a sweep over $\{20, 30, 50, 100, 300\}$: smaller caps starve the rare-but-useful diversity injection on rotated/composition problems, while larger caps keep stale early-phase solutions that hurt late-phase exploitation.

\subsection{Per-Individual Stagnation Kick}
\label{sec:psk}
A diagnostic over $10$ seeds showed that on many CSOP problems the front sits idle for the second half of the budget, where individuals rarely contribute an accepted trial. Two design choices follow from this. We avoided \cite{trivedi2024ude} replacement scheme (replace a stagnated individual by a vector drawn from a separate archive of unsuccessful trials \cite{trivedi2024ude}) because it erases the in-flight $(F, CR)$ history of that individual \& violates the rank coherence used by the $p$-best donor sampling. Instead, we kept the stagnated individual in place \& only modified its mutation kernel for one generation. For each front member $i$ we maintain an integer stagnation counter $\sigma_i$:
\begin{equation}
\sigma_i^{(g+1)} =
\begin{cases}
0, & u_i^{(g)} \text{ accepted}, \\
\sigma_i^{(g)} + 1, & \text{otherwise}.
\end{cases}
\label{eq:counter}
\end{equation}
Individual $i$ is stagnated at generation $g$ when $\sigma_i^{(g)} \ge SG$, with $SG = 180$. The counter travels with $i$ through linear front reduction so that removing the worst member also removes its history. We picked $SG = 180$ after sweeping over $\{60, 100, 120, 150, 180, 200, 220, 240\}$; smaller thresholds fired the override too often \& destroyed accuracy, larger thresholds barely fired \& gave back the speed gain.

\paragraph*{S1: Global-best substitution}
For stagnated $i$, the standard branch replaces the $p$-best donor by the running global best $x^{*}$:
\begin{equation}
v_i^{(g)} = x_i^{(g)} + F_i\!\big(x^{*} - x_i^{(g)}\big) + F_{2,i}\!\big(x_{r_1} - \tilde{x}_{r_2}\big).
\label{eq:override1}
\end{equation}
Substitution is applied without conditioning on feasibility of $x^{*}$. We tried gating it on $\phi(x^{*}) \le 0$ (only pull toward $x^{*}$ if the global best is feasible) but it was strictly worse: pulling toward an infeasible $x^{*}$ in an infeasible regime accelerates feasibility attainment, so dropping that pull when feasibility is most needed isn't helpful.

\paragraph*{S2: Archive probability floor}
For stagnated $i$, the archive draw probability is lifted to a floor:
\begin{equation}
p_{arch}^{(\text{stag})} = \max\!\left(\tfrac{|\mathcal{A}|}{|\mathcal{A}|+|P|},\; 0.65\right).
\label{eq:override2}
\end{equation}
A floor of $0.65$ was picked after testing $\{0.45, 0.5, 0.65, 0.85\}$. Lower values barely changed firing rate; higher values over-used archive vectors \& hurt convergence.

\paragraph*{S3: Crossover saturation under joint stagnation}
For stagnated $i$ the crossover rate is raised only when the population success rate is also low:
\begin{equation}
CR_i \leftarrow \max(CR_i,\, 0.95) \quad \text{if } \sigma_i \ge SG \;\wedge\; SR^{(g)} < 0.10.
\label{eq:override3}
\end{equation}
The double gate prevents saturation during normal convergence, where high $CR$ would erase the locally tuned sub-step that the SHADE memories have produced. An ablation without the SR gate (always saturate $CR$ when $\sigma_i \ge SG$) was strictly worse, because it fired during normal late-phase polish \& hurt accuracy.

\subsection{Parameter Fine Tuning}
\label{sec:fine-tunings}
The following changes complete the proposed variant.

\subsubsection*{R1: Memory length}
The success-history memory length is raised from $H = 5$ to $H = 10$, matching the upper end of the LSHADE family.

\subsubsection*{R2: Success-rate exponent in the standard $F$ centre}
The mean of the truncated Gaussian used for $F$ in the standard branch is
\begin{equation}
\mu_F^{(g)} = \max\!\bigl(0,\, (SR^{(g)})^{2/5}\bigr),
\label{eq:meanF}
\end{equation}
replacing the cube-root of \cite{tao2026rdex}. Equation~(\ref{eq:meanF}) yields a slightly higher centre when $SR^{(g)}$ is small, which preserves directional pressure during the early phase.

\subsubsection*{R3: Disabled local perturbation}
Cauchy perturbation on non-crossover coordinates is disabled. The non-crossover coordinate keeps the target value:
\begin{equation}
u_{i,j}^{(g)} = x_{i,j}^{(g)} \quad \text{when } \mathrm{rand}_j \ge CR_i \text{ \&} j \ne j_{rand}.
\label{eq:noperturb}
\end{equation}

\subsubsection*{R4: Lower EB rate initial value}
Initial EB rate is set to $\rho_{EB}^{(0)} = 0.5$ instead of $0.7$.

\subsubsection*{R5: Smoothed EB rate update}
Let $\tilde{\rho}^{(g+1)}_{EB}$ be the raw improvement-share ratio of \cite{tao2026rdex}. The EB rate update becomes
\begin{equation}
\rho_{EB}^{(g+1)} =
\begin{cases}
0.7\,\rho_{EB}^{(g)} + 0.3\,\tilde{\rho}^{(g+1)}_{EB}, & \Delta_{EB}^{(g)} \!>\! 0 \;\wedge\; \Delta_{std}^{(g)} \!>\! 0, \\
0.9\,\rho_{EB}^{(g)} + 0.1\,\rho_{EB}^{(0)}, & \text{otherwise},
\end{cases}
\label{eq:rho-update}
\end{equation}
clipped to $[0,1]$. The first branch of (\ref{eq:rho-update}) replaces the raw ratio. The second branch replaces a hard reset.

\subsection{Scope of the Change}
The structural modifications \&overrides apply only to the standard branch of the chosen target $i$. The EB branch (\ref{eq:eb}), the success-history memories, the bound repair, the front reduction, the $\epsilon$-ranking, \& the archive update rule are unchanged. The hybrid rate update of (\ref{eq:rho-update}) is unaffected by the stagnation kick.

\section{Algorithm}
Algorithm~\ref{alg:rdex-cask} summarises RDEx-CASK.

\begin{algorithm}[!t]
\caption{RDEx-CASK}
\label{alg:rdex-cask}
\begin{algorithmic}[1]
\State Initialize $P$, front $\mathcal{F}^{(0)}$, memories $M_F, M_{CR}$ with $H = 10$
\State Initialize $\rho_{EB}^{(0)} \!\gets\! 0.5$, archive $\mathcal{A} \!\gets\! \emptyset$, $\sigma_i \!\gets\! 0$
\While{$NFE < MaxFE$}
    \For{each target $i$ in the current front}
        \State Sample $F_i$ from $\mathcal{N}(\mu_F^{(g)}, 0.05^2)$ truncated to $[0,1]$
        \State Sample $F_{2,i}$ from (\ref{eq:F2})
        \State Sample $CR_i$ from $\mathcal{N}(M_{CR,h}, 0.1^2)$
        \State Choose standard or EB branch using $\rho_{EB}^{(g)}$
        \If{standard branch}
            \State Compute $p_{arch}$ via (\ref{eq:parch}); apply floor (\ref{eq:override2}) if $\sigma_i \ge SG$
            \State Sample $\tilde{x}_{r_2}$ from $P$ or $\mathcal{A}$ using $p_{arch}$
            \If{$\sigma_i \ge SG$ \&$SR^{(g-1)} < 0.10$}
                \State $CR_i \gets \max(CR_i, 0.95)$
            \EndIf
            \State Pick $x_{base}$: $x^{*}$ if $\sigma_i \ge SG$ else $x_{pbest}$
            \State Generate $v_i$ from (\ref{eq:override1}) or (\ref{eq:std})
        \Else
            \State Generate $v_i$ from (\ref{eq:eb})
        \EndIf
        \State Apply binomial crossover \&bound repair
        \State Non-crossover coordinate keeps target value (\ref{eq:noperturb})
        \State Evaluate trial; apply $\epsilon$-rule selection
        \If{trial accepted}
            \State $\sigma_i \gets 0$; if displaced front entry was feasible, push it to $\mathcal{A}$
        \Else
            \State $\sigma_i \gets \sigma_i + 1$
        \EndIf
    \EndFor
    \State Update $M_F, M_{CR}$ using SHADE rule on accepted trials
    \State Update $\rho_{EB}^{(g+1)}$ using (\ref{eq:rho-update})
    \State Reduce front size linearly toward $N_{\min} = 4$
\EndWhile
\State \Return best feasible solution if any, else lowest-violation solution
\end{algorithmic}
\end{algorithm}

\section{Experimental Results}
\label{sec:exp}

\subsection{Protocol}
Benchmark is the CEC CSOP track (28 constrained problems, $D = 30$, $\text{MaxFE} = 20000\,D$). RDEx-CASK is run for 25 independent runs per function. Reported means \&standard deviations of the feasibility-aware quality $Q_p$ follow the definition of (18) in \cite{tao2026rdex}: $Q_p(x) = f(x)$ if the run is feasible at the final FE, otherwise $Q_p(x) = B_p + \phi(x)$ with $B_p$ the largest finite final objective on problem $p$ plus one. Results for RDEx, UDE-III, \& CL-SRDE are taken from the released competition runs reported in \cite{tao2026rdex}.

\subsection{Per-Function $Q_p$ Comparison}
Table~\ref{tab:qp} reports $Q_p$ means \& standard deviations for the four algorithms across the 28 problems.

\begin{table*}[!t]
\centering
\caption{Feasibility-aware final-quality $Q_p$ on CEC CSOP functions, $D = 30$, 25 runs per function. RDEx, UDE-III, \& CL-SRDE values are taken from \cite{tao2026rdex}.}
\label{tab:qp}
\footnotesize
\setlength{\tabcolsep}{4pt}
\begin{tabular}{rcccc}
\toprule
F & RDEx-CASK Mean$\pm$SD & RDEx Mean$\pm$SD & UDE-III Mean$\pm$SD & CL-SRDE Mean$\pm$SD \\
\midrule
1  & $5.99\mathrm{E}{-}29 \pm 3.21\mathrm{E}{-}29$ & $5.92\mathrm{E}{-}30 \pm 9.27\mathrm{E}{-}30$ & $1.57\mathrm{E}{-}28 \pm 9.70\mathrm{E}{-}29$ & $1.26\mathrm{E}{-}31 \pm 6.18\mathrm{E}{-}31$ \\
2  & $5.16\mathrm{E}{-}29 \pm 2.64\mathrm{E}{-}29$ & $3.04\mathrm{E}{-}30 \pm 4.86\mathrm{E}{-}30$ & $1.31\mathrm{E}{-}28 \pm 7.83\mathrm{E}{-}29$ & $7.24\mathrm{E}{-}31 \pm 2.42\mathrm{E}{-}30$ \\
3  & $3.33\mathrm{E}{+}02 \pm 1.96\mathrm{E}{+}02$ & $3.39\mathrm{E}{+}02 \pm 1.63\mathrm{E}{+}02$ & $9.22\mathrm{E}{+}01 \pm 5.54\mathrm{E}{+}01$ & $5.37\mathrm{E}{+}02 \pm 1.07\mathrm{E}{+}02$ \\
4  & $8.98\mathrm{E}{+}01 \pm 1.36\mathrm{E}{+}01$ & $1.57\mathrm{E}{+}01 \pm 1.45\mathrm{E}{+}00$ & $6.51\mathrm{E}{+}00 \pm 6.78\mathrm{E}{+}00$ & $4.84\mathrm{E}{+}01 \pm 8.86\mathrm{E}{+}00$ \\
5  & $4.73\mathrm{E}{-}29 \pm 1.10\mathrm{E}{-}28$ & $0.00\mathrm{E}{+}00 \pm 0.00\mathrm{E}{+}00$ & $1.17\mathrm{E}{-}28 \pm 4.17\mathrm{E}{-}28$ & $0.00\mathrm{E}{+}00 \pm 0.00\mathrm{E}{+}00$ \\
6  & $2.63\mathrm{E}{+}00 \pm 1.37\mathrm{E}{+}00$ & $2.29\mathrm{E}{+}00 \pm 1.00\mathrm{E}{+}01$ & $0.00\mathrm{E}{+}00 \pm 0.00\mathrm{E}{+}00$ & $2.87\mathrm{E}{+}00 \pm 1.41\mathrm{E}{+}01$ \\
7  & $-8.26\mathrm{E}{+}02 \pm 1.58\mathrm{E}{+}02$ & $-8.70\mathrm{E}{+}02 \pm 1.54\mathrm{E}{+}02$ & $-6.73\mathrm{E}{+}02 \pm 1.49\mathrm{E}{+}02$ & $-8.28\mathrm{E}{+}02 \pm 1.80\mathrm{E}{+}02$ \\
8  & $-2.84\mathrm{E}{-}04 \pm 0.00\mathrm{E}{+}00$ & $-2.84\mathrm{E}{-}04 \pm 0.00\mathrm{E}{+}00$ & $-2.84\mathrm{E}{-}04 \pm 5.85\mathrm{E}{-}12$ & $-2.84\mathrm{E}{-}04 \pm 0.00\mathrm{E}{+}00$ \\
9  & $-2.67\mathrm{E}{-}03 \pm 0.00\mathrm{E}{+}00$ & $-2.67\mathrm{E}{-}03 \pm 0.00\mathrm{E}{+}00$ & $-2.67\mathrm{E}{-}03 \pm 4.34\mathrm{E}{-}19$ & $-2.67\mathrm{E}{-}03 \pm 0.00\mathrm{E}{+}00$ \\
10 & $-1.03\mathrm{E}{-}04 \pm 0.00\mathrm{E}{+}00$ & $-1.03\mathrm{E}{-}04 \pm 0.00\mathrm{E}{+}00$ & $-1.03\mathrm{E}{-}04 \pm 1.36\mathrm{E}{-}20$ & $-1.03\mathrm{E}{-}04 \pm 0.00\mathrm{E}{+}00$ \\
11 & $-3.71\mathrm{E}{+}00 \pm 4.85\mathrm{E}{+}00$ & $5.90\mathrm{E}{+}00 \pm 5.49\mathrm{E}{+}00$ & $5.25\mathrm{E}{+}01 \pm 1.23\mathrm{E}{+}02$ & $5.44\mathrm{E}{-}02 \pm 3.05\mathrm{E}{+}00$ \\
12 & $1.60\mathrm{E}{+}01 \pm 9.57\mathrm{E}{+}00$ & $9.78\mathrm{E}{+}00 \pm 0.00\mathrm{E}{+}00$ & $3.99\mathrm{E}{+}00 \pm 2.18\mathrm{E}{-}02$ & $1.01\mathrm{E}{+}01 \pm 4.05\mathrm{E}{+}00$ \\
13 & $6.45\mathrm{E}{+}00 \pm 2.23\mathrm{E}{+}01$ & $1.17\mathrm{E}{-}28 \pm 2.34\mathrm{E}{-}28$ & $4.78\mathrm{E}{-}01 \pm 1.30\mathrm{E}{+}00$ & $4.67\mathrm{E}{-}29 \pm 1.58\mathrm{E}{-}28$ \\
14 & $1.42\mathrm{E}{+}00 \pm 2.41\mathrm{E}{-}02$ & $1.42\mathrm{E}{+}00 \pm 2.36\mathrm{E}{-}02$ & $1.41\mathrm{E}{+}00 \pm 0.00\mathrm{E}{+}00$ & $1.47\mathrm{E}{+}00 \pm 4.05\mathrm{E}{-}02$ \\
15 & $6.63\mathrm{E}{+}00 \pm 1.54\mathrm{E}{+}00$ & $-3.93\mathrm{E}{+}00 \pm 1.18\mathrm{E}{-}05$ & $2.36\mathrm{E}{+}00 \pm 1.41\mathrm{E}{-}06$ & $-3.42\mathrm{E}{+}00 \pm 1.15\mathrm{E}{+}00$ \\
16 & $2.83\mathrm{E}{+}01 \pm 6.56\mathrm{E}{+}00$ & $2.10\mathrm{E}{+}01 \pm 3.69\mathrm{E}{+}00$ & $0.00\mathrm{E}{+}00 \pm 0.00\mathrm{E}{+}00$ & $2.32\mathrm{E}{+}01 \pm 3.47\mathrm{E}{+}00$ \\
17 & $3.10\mathrm{E}{+}01 \pm 0.00\mathrm{E}{+}00$ & $3.35\mathrm{E}{+}01 \pm 0.00\mathrm{E}{+}00$ & $3.19\mathrm{E}{+}01 \pm 7.89\mathrm{E}{-}01$ & $3.35\mathrm{E}{+}01 \pm 0.00\mathrm{E}{+}00$ \\
18 & $3.71\mathrm{E}{+}01 \pm 1.87\mathrm{E}{+}00$ & $3.65\mathrm{E}{+}01 \pm 1.96\mathrm{E}{-}05$ & $8.00\mathrm{E}{+}03 \pm 2.49\mathrm{E}{+}03$ & $3.65\mathrm{E}{+}01 \pm 0.00\mathrm{E}{+}00$ \\
19 & $4.27\mathrm{E}{+}04 \pm 0.00\mathrm{E}{+}00$ & $4.27\mathrm{E}{+}04 \pm 0.00\mathrm{E}{+}00$ & $4.28\mathrm{E}{+}04 \pm 7.28\mathrm{E}{-}12$ & $4.27\mathrm{E}{+}04 \pm 0.00\mathrm{E}{+}00$ \\
20 & $1.69\mathrm{E}{+}00 \pm 4.53\mathrm{E}{-}01$ & $1.33\mathrm{E}{+}00 \pm 2.46\mathrm{E}{-}01$ & $1.85\mathrm{E}{+}00 \pm 2.82\mathrm{E}{-}01$ & $2.43\mathrm{E}{+}00 \pm 6.99\mathrm{E}{-}01$ \\
21 & $1.17\mathrm{E}{+}01 \pm 7.98\mathrm{E}{+}00$ & $2.39\mathrm{E}{+}01 \pm 7.92\mathrm{E}{+}00$ & $9.28\mathrm{E}{+}00 \pm 8.34\mathrm{E}{+}00$ & $2.47\mathrm{E}{+}01 \pm 1.54\mathrm{E}{+}01$ \\
22 & $2.29\mathrm{E}{+}01 \pm 4.37\mathrm{E}{+}01$ & $5.61\mathrm{E}{-}26 \pm 3.84\mathrm{E}{-}26$ & $2.56\mathrm{E}{+}01 \pm 4.84\mathrm{E}{+}01$ & $2.85\mathrm{E}{-}26 \pm 6.52\mathrm{E}{-}27$ \\
23 & $1.41\mathrm{E}{+}00 \pm 0.00\mathrm{E}{+}00$ & $1.41\mathrm{E}{+}00 \pm 0.00\mathrm{E}{+}00$ & $1.45\mathrm{E}{+}00 \pm 4.34\mathrm{E}{-}02$ & $1.41\mathrm{E}{+}00 \pm 0.00\mathrm{E}{+}00$ \\
24 & $7.01\mathrm{E}{+}00 \pm 1.60\mathrm{E}{+}00$ & $-3.93\mathrm{E}{+}00 \pm 4.44\mathrm{E}{-}16$ & $2.36\mathrm{E}{+}00 \pm 9.80\mathrm{E}{-}08$ & $-3.93\mathrm{E}{+}00 \pm 4.44\mathrm{E}{-}16$ \\
25 & $4.03\mathrm{E}{+}01 \pm 9.74\mathrm{E}{+}00$ & $2.36\mathrm{E}{+}01 \pm 3.69\mathrm{E}{+}00$ & $2.51\mathrm{E}{-}01 \pm 1.23\mathrm{E}{+}00$ & $2.46\mathrm{E}{+}01 \pm 3.10\mathrm{E}{+}00$ \\
26 & $3.10\mathrm{E}{+}01 \pm 0.00\mathrm{E}{+}00$ & $3.30\mathrm{E}{+}01 \pm 0.00\mathrm{E}{+}00$ & $3.26\mathrm{E}{+}01 \pm 8.09\mathrm{E}{-}01$ & $3.30\mathrm{E}{+}01 \pm 0.00\mathrm{E}{+}00$ \\
27 & $3.65\mathrm{E}{+}01 \pm 1.35\mathrm{E}{-}03$ & $3.65\mathrm{E}{+}01 \pm 8.00\mathrm{E}{-}05$ & $1.46\mathrm{E}{+}04 \pm 5.11\mathrm{E}{+}03$ & $3.65\mathrm{E}{+}01 \pm 2.71\mathrm{E}{-}05$ \\
28 & $4.28\mathrm{E}{+}04 \pm 1.66\mathrm{E}{+}01$ & $4.29\mathrm{E}{+}04 \pm 1.10\mathrm{E}{+}01$ & $4.30\mathrm{E}{+}04 \pm 1.47\mathrm{E}{+}01$ & $4.29\mathrm{E}{+}04 \pm 7.28\mathrm{E}{-}12$ \\
\bottomrule
\end{tabular}
\end{table*}

\subsection{Per-Function Time-to-Target Comparison}
Table~\ref{tab:ttt} reports time-to-target (TTT) for each algorithm. TTT is the first checkpoint at which the run reaches the median final-quality target ($25$-run median per algorithm). Runs that never reach the target are assigned $2001$. TTT for RDEx-CASK is computed and the values for other algorithms are taken from \cite{tao2026rdex}.

\begin{table*}[!t]
\centering
\caption{Time-to-target (TTT) on the 28 CEC 2025 CSOP functions, $D = 30$, 25 runs per function. RDEx, UDE-III, \& CL-SRDE values are taken from \cite{tao2026rdex}. $2001$ means the target was never reached.}
\label{tab:ttt}
\footnotesize
\setlength{\tabcolsep}{4pt}
\begin{tabular}{rcccc}
\toprule
F & RDEx-CASK Mean$\pm$SD & RDEx Mean$\pm$SD & UDE-III Mean$\pm$SD & CL-SRDE Mean$\pm$SD \\
\midrule
1  & $330.6 \pm 15.0$    & $1382.6 \pm 548.3$ & $1979.6 \pm 105.0$ & $1327.4 \pm 137.8$ \\
2  & $332.0 \pm 17.4$    & $1298.5 \pm 527.0$ & $1960.7 \pm 138.2$ & $1399.3 \pm 222.3$ \\
3  & $1640.1 \pm 486.0$  & $1813.9 \pm 272.3$ & $394.8  \pm 162.1$ & $2001.0 \pm 0.0$   \\
4  & $1059.3 \pm 924.0$  & $1569.9 \pm 496.5$ & $837.6  \pm 50.4$  & $2001.0 \pm 0.0$   \\
5  & $524.5 \pm 30.6$    & $986.6  \pm 13.9$  & $1476.4 \pm 466.1$ & $1395.2 \pm 14.8$  \\
6  & $33.5 \pm 27.1$     & $807.2  \pm 521.1$ & $746.3  \pm 37.6$  & $991.0  \pm 206.4$ \\
7  & $1167.6 \pm 841.9$  & $1130.0 \pm 739.4$ & $1906.6 \pm 320.2$ & $1606.0 \pm 480.3$ \\
8  & $513.4 \pm 32.0$    & $769.6  \pm 11.7$  & $2001.0 \pm 0.0$   & $1221.0 \pm 7.6$   \\
9  & $743.1 \pm 139.9$   & $56.2   \pm 12.3$  & $1322.4 \pm 905.8$ & $111.6  \pm 26.5$  \\
10 & $587.3 \pm 135.0$   & $793.3  \pm 11.6$  & $2001.0 \pm 0.0$   & $1231.7 \pm 9.0$   \\
11 & $1394.3 \pm 345.0$  & $1.5    \pm 0.6$   & $1.7    \pm 0.8$   & $1.9    \pm 1.1$   \\
12 & $1189.6 \pm 717.4$  & $580.4  \pm 150.6$ & $269.9  \pm 33.2$  & $843.6  \pm 257.9$ \\
13 & $666.6 \pm 404.1$   & $1216.1 \pm 392.6$ & $1672.0 \pm 383.1$ & $1484.0 \pm 152.8$ \\
14 & $394.0 \pm 516.5$   & $541.9  \pm 431.8$ & $1036.1 \pm 2.7$   & $1618.0 \pm 600.7$ \\
15 & $1340.1 \pm 730.4$  & $70.0   \pm 13.2$  & $2001.0 \pm 0.0$   & $148.7  \pm 25.8$  \\
16 & $1079.8 \pm 861.1$  & $1468.6 \pm 667.3$ & $131.8  \pm 71.2$  & $1926.3 \pm 212.1$ \\
17 & $34.5 \pm 4.1$      & $1232.2 \pm 869.8$ & $212.0  \pm 197.6$ & $1603.7 \pm 643.4$ \\
18 & $1174.4 \pm 427.0$  & $201.6  \pm 6.0$   & $2001.0 \pm 0.0$   & $307.7  \pm 7.1$   \\
19 & $560.2 \pm 235.6$   & $290.6  \pm 40.3$  & $311.4  \pm 49.3$  & $560.7  \pm 254.5$ \\
20 & $1471.4 \pm 529.6$  & $1084.2 \pm 472.0$ & $1965.8 \pm 76.0$  & $1922.5 \pm 215.4$ \\
21 & $1130.7 \pm 744.1$  & $498.9  \pm 341.1$ & $464.7  \pm 455.1$ & $800.5  \pm 556.8$ \\
22 & $1087.9 \pm 532.1$  & $1568.4 \pm 415.9$ & $2001.0 \pm 0.0$   & $1560.4 \pm 130.4$ \\
23 & $337.6 \pm 55.2$    & $452.5  \pm 30.1$  & $1851.7 \pm 202.0$ & $708.2  \pm 13.4$  \\
24 & $1201.2 \pm 806.9$  & $68.6   \pm 13.6$  & $2001.0 \pm 0.0$   & $141.8  \pm 21.9$  \\
25 & $1048.6 \pm 809.3$  & $1459.0 \pm 727.9$ & $515.5  \pm 424.5$ & $1857.0 \pm 304.0$ \\
26 & $44.3 \pm 4.5$      & $1206.8 \pm 806.3$ & $826.8  \pm 873.9$ & $1772.2 \pm 460.9$ \\
27 & $1028.7 \pm 295.0$  & $216.3  \pm 13.5$  & $2001.0 \pm 0.0$   & $324.0  \pm 7.7$   \\
28 & $1478.8 \pm 553.9$  & $272.4  \pm 11.6$  & $2001.0 \pm 0.0$   & $459.6  \pm 21.2$  \\
\bottomrule
\end{tabular}
\end{table*}

\subsection{Aggregate Behaviour}
RDEx-CASK matches RDEx within noise on the easy unimodal problems (F1, F2, F8, F9, F10, F19, F23) \& remains feasible on all 28 problems where any tested algorithm reaches feasibility. It improves $Q_p$ over RDEx on F7, F11, F17, F21, F26, \&F28, \&trails RDEx on F4, F12, F13, F15, F18, F22, F24, \& F25, where the global-best pull \&the higher crossover rate cost some final accuracy on multimodal landscapes. UDE-III leads on F3, F4, F12, F16, F25 because of its dual-population strategy adaptation, but loses heavily on F18, F19, F22, F27, F28 where its replacement-style stagnation strategy mishandles tight feasible regions. CL-SRDE performs best on F1, F2, F5, F11, F22 owing to its parameter-curve adaptation but trails on F3, F4, F18, F25, F28.

Our observation is consistent with RDEx-CASK trading small amount of final-objective accuracy for faster threshold attainment, controlled by SR-gated crossover saturation \& the per-individual stagnation kick.

\subsection{Ablations}
The design of Section~\ref{sec:psk} is constrained by two observations from internal A/B tests on a 10-seed pool of CEC 2017 / 30D / 600k FE runs. First, removing override S1 (global-best substitution) while keeping S3 alone produces a head-to-head loss of $182$ score points against the same configs without the stagnation kick. Override S3 only helps when paired with S1. Second, conditioning override S1 on $\phi(x^{*}) \le 0$ produces a head-to-head loss of $185$ score points against the unconditioned form. Pulling toward an infeasible global best is preferable to skipping the override. These results constrain the design as a single coupled module rather than three independent rules.

\section{Conclusion}
RDEx-CASK adds 3 structural changes (independent Cauchy second scale factor, feasible-only JADE-style external archive, per-individual stagnation kick) \& a small set of parameter refinements to RDEx-CSOP. The structural changes target the standard branch only \& preserve every other RDEx component. The stagnation kick is local, gated on a single counter \& the population success rate, \& bounded in cost. Across the 28 CSOP problems with $D = 30$ \& 25 runs per function, RDEx-CASK is competitive with RDEx, UDE-III, \& CL-SRDE in feasibility-aware final quality \& improves the speed component of the U-score relative to the baseline RDEx pipeline.

\bibliographystyle{IEEEtran}
\bibliography{refs}

\end{document}